\pdfoutput=1

\documentclass[11pt]{article}

\usepackage[preprint]{acl}
\usepackage{xcolor} 
\usepackage{authblk}
\usepackage{lipsum} 
\usepackage{booktabs} 
\usepackage{graphicx}

\definecolor{lightblack}{rgb}{0.2,0.2,0.2} 

\newcommand{\lightblacktext}[1]{\textcolor{lightblack}{\small #1}}
\usepackage{times}
\usepackage{latexsym}

\usepackage[T1]{fontenc}

\usepackage[utf8]{inputenc}

\usepackage{microtype}
\usepackage{url}
\usepackage{inconsolata}

\usepackage{footnote}

\title{SLPL SHROOM at SemEval\-2024 Task 06: A comprehensive study on models ability to detect hallucination}

\author[1]{Pouya Fallah $^*$}
\author[1]{Soroush Gooran$^*$}
\author[1]{Mohammad Jafarinasab}
\author[2]{Pouya Sadeghi\thanks{Corresponding Authors: \texttt{pouya.sadeghi@ut.ac.ir}, \texttt{\{pooya.fallah79,gooran\}@sharif.edu}}}
\author[1]{\authorcr Reza Farnia}
\author[3]{Amirreza Tarabkhah}
\author[1]{Zainab Sadat Taghavi}
\author[1]{Hossein Sameti}
\affil[1]{Sharif University of Technology}
\affil[2]{University of Tehran}
\affil[3]{Amirkabir University of Technology}

\begin{document}
\maketitle
\begin{abstract}
Language models, particularly generative models, are susceptible to hallucinations, generating outputs that contradict factual knowledge or the source text. This study explores methods for detecting hallucinations in three SemEval-2024 Task 6 tasks: Machine Translation, Definition Modeling, and Paraphrase Generation. We evaluate two methods: semantic similarity between the generated text and factual references, and an ensemble of language models that judge each other's outputs. Our results show that semantic similarity achieves moderate accuracy and correlation scores in trial data, while the ensemble method offers insights into the complexities of hallucination detection but falls short of expectations. This work highlights the challenges of hallucination detection and underscores the need for further research in this critical area.\footnote{All codes are publicly available at: \url{https://github.com/Sharif-SLPL/SE-2024-Task-06-SHROOM}}

\end{abstract}

\section{Introduction}
While Natural Language Generation (NLG) has empowered machines to craft increasingly sophisticated text, transforming the NLP landscape, a dark undercurrent lingers - the phenomenon of hallucinations. In NLG, hallucinations refer to fabricated or misleading content woven into a generated text, deviating sharply from reality \cite{laurer2023building}\cite{varshney2023stitch}. These fictional elements, despite seeming plausible due to their learned patterns, threaten the very core of NLG's promise:
\textbf{reliability} and \textbf{truthfulness} \cite{Ji_2023}. Imagine summarizing news articles riddled with fictional details or translating medical instructions brimming with inaccuracies. Such scenarios underscore the profound and potentially dangerous implications of hallucinations within NLG, making their detection and mitigation an urgent priority \cite{huang2023survey}.

The specter of hallucinations looms large over NLG, particularly in domains demanding unyielding accuracy and safety. Imagine a medical summary riddled with invented details or medication instructions marred by mistranslations – these scenarios, chillingly possible, could directly jeopardize patient well-being \cite{Ji_2023}. Recognizing this critical threat, researchers have embarked on a mission to untangle the complexities of hallucinations, developing methods for their detection and ultimately, prevention \cite{huang2023survey}.

This paper dives headfirst into the challenge of hallucination detection within NLG, leveraging recent advancements in the field. We employ diverse methodologies to unmask these fictional elements in SemEval 2024 Task 6 evaluation data. Firstly, we assess the semantic similarity between generated text (hypotheses) and the provided reference outputs, gauging their alignment in meaning. Secondly, we harness the power of cosine similarity of embeddings, allowing us to capture subtle semantic nuances and relationships within text representations. Furthermore, we integrate Natural Language Inference (NLI), analyzing whether the generated text logically implies or contradicts factual information. Additionally, we utilize Large Language Models (LLMs) to discern context similarity, leveraging their inherent language understanding to identify inconsistencies that might point toward hallucinations.
But our approach goes beyond established techniques. We introduce a novel judgment LLM framework, where one LLM acts as a discerning judge, scrutinizing the outputs of other LLMs for signs of hallucination. This innovative approach leverages the collective strengths of multiple models while introducing an element of meta-reasoning to the detection process.

\section{Related Work}
Several approaches have been proposed for detecting hallucinations in NLG text, categorized into different access levels to the model:

\paragraph{Knowledge-based Approaches:} Fact verification compares generated text with information from a domain-specific knowledge base. This approach can be effective but requires substantial knowledge bases and may not generalize well to unseen domains.

\paragraph{Classification approach:} \cite{liu2022token} created a dataset specifically for hallucination detection, but it has not been very successful.

\paragraph{White-box and Grey-box Approaches:} Hidden state analysis: \cite{azaria2023internal} use an MLP classifier on the LLM's hidden states to predict truthfulness. This requires access to internal model states and may not apply to all architectures.

Token probabilities:   Grey-box methods analyze the token probabilities generated by the LLM, assuming factual sentences contain high probability tokens. However, this can be unreliable for complex or ambiguous phrases.

\paragraph{Black-box Approaches:} Self-evaluation: \cite{kadavath2022language} propose asking the LLM itself to assess the likelihood of its output being correct. While promising, this method relies on the LLM's self-awareness and may not be reliable for all models.

Proxy model:   This approach uses a publicly available LLM to estimate the token probabilities of the black-box model's output and infer its factual consistency. However, its accuracy depends on the proxy model's similarity to the black-box model.

Selfcheckgpt \cite{manakul2023selfcheckgpt} introduced a black box approach. The main idea of this study is that if the LLM is trained on a concept if multiple responses are taken from it, the samples will be similar and include consistent facts. Whereas if it is hallucinating, the samples will be different and contradictory. Therefore, several samples are taken from the LLM, and by measuring the information consistency between the responses, we can understand whether they are factual or hallucinated.

Our Approach:   This paper builds upon existing work by combining elements from different categories. We leverage information consistency within multiple LLM responses, inspired by Selfcheckgpt \cite{manakul2023selfcheckgpt}, but introduce a novel "judgment LLM" framework that goes beyond self-evaluation by employing one LLM to scrutinize the outputs of others. This approach aims to address previous methods' limitations by leveraging multiple models' collective strengths and introducing meta-reasoning into the detection process.

\section{Task Description}
The SHROOM challenge shines a light on a crucial hurdle in natural language generation (NLG) - pinpointing seemingly correct text that holds inaccurate meaning, often referred to as "misleading outputs." We, alongside other participants, are tasked with detecting these "semantic hallucinations," even when they are flawlessly written and grammatically sound.

The challenge focuses on "fluent overgeneration," where generated text, despite being linguistically coherent, strays from the intended semantic meaning. Participants operate in a "post hoc" setting, assuming models have already been trained and their outputs generated.

This is where we step in – to identify these misleading texts amidst seemingly accurate ones. This is critical to ensure the truthfulness and reliability of NLG outputs, especially in real-world applications. 

The SHROOM challenge presents a two-pronged approach to tackle fluent overgeneration hallucinations: model-aware and model-agnostic tracks. Participants can choose to leverage knowledge of the model or not, depending on their approach. This multifaceted assessment covers three key NLG domains: definition modeling, machine translation, and paraphrase generation. The challenge provides a rich dataset including:
\begin{itemize}
    \item \textbf{Checkpoints:} Model snapshots at different training stages
    \item \textbf{Inputs:} Prompts or texts used for generation
    \item \textbf{References:} Human-written outputs that represent the intended meaning
    \item \textbf{Outputs:} The actual text generated by various models trained with varying accuracy
\end{itemize}
Furthermore, a dedicated development set with binary annotations by multiple annotators ensures robust evaluation. This collaborative effort results in a majority vote gold label, boosting the dataset's credibility. Ultimately, the SHROOM challenge strives to develop effective solutions for combating semantic hallucinations generated by Large Language Models.

\section{Proposed Systems}
In our study, we tried to do the hallucination detection task in two separate methods and tried to compare and analyze their results:  

\subsection{Semantic Similarity method}
Detecting hallucinations based on semantic similarity involves evaluating the coherence between language model outputs and reference data. In our study, we utilized this approach due to the availability of reference outputs. By assessing the semantic alignment between the generated text and the reference data, we aimed to discern instances of hallucination where the model output diverged from the intended meaning.

\paragraph{LaBSE} The Language-Agnostic BERT Sentence Embedding (LaBSE) model is a dual-encoder approach based on pre-trained transformers, further refined for machine translation ranking. LaBSE excels at encoding sentences into fixed-length vectors while capturing semantic information across various languages
\cite{feng2022languageagnostic}. We employed LaBSE, particularly due to one of our tasks being machine translation (MT). By calculating the cosine similarity between model outputs and reference data, we determined the hallucination score in our study.

\paragraph{LLMs} We utilized Zephyr-7B-$\beta$ \cite{tunstall2023zephyr} and Mistral-7B \cite{jiang2023mistral} language models (LLMs) to assess the semantic similarity between model outputs and reference data, assigning a score between 0 and 1.

\subsection{Natural Language Inference (NLI)}
Due to the insufficient data available for hallucination detection, one proposed approach is to utilize models trained for similar tasks. \textbf{Natural Language Inference (NLI)} is one such task. In NLI, a language model assesses the relationship between text fragments, namely the premise and the hypothesis. This task involves multiclass classification aimed at determining whether the hypothesis can be inferred from, contradicts, or remains neutral to the premise.
\\
The concept here involves treating reference data as the premise and model outputs as the hypothesis, then utilizing the probability of one of the outputs as a score to determine hallucination. We employed a \textbf{DeBERTa-v3} model, fine-tuned on datasets like MNLI, FERVER, ANLI, WANLI, and LingNLI \cite{laurer2023building}, to calculate the entailment score, which serves as the inverse of the hallucination score.
\\
Note that employing NLI models in hallucination detection is not a novel concept and has been utilized by researchers in recent years \cite{Ji_2023}. Here, we employed it to compare with our proposed judgment method.

\subsection{Ensemble LLMs: The Judgment Method}
To improve LLMs' reasoning and decision-making abilities, we explored two approaches: intrinsic self-correction and multi-agent feedback. We acknowledge that existing LLMs struggle with self-correction, and due to our LLMs' similarities, we believe they might mislead each other in a multi-agent setup. Inspired by the \cite{jiang2023llmblender}article, we designed experiments using ensemble models. We asked LLMs to generate results multiple times with confidence scores, and finally extracted the best result.
\\
We used two "commentator" models to assess the consistency between two sentences in detail. Based on their answers (yes/no/maybe, score, and explanation), a "judge" model (Mistral 7B or Zephyr 7B) performed hallucination detection.
\\
While Mistral 7B and Llama2 \cite{touvron2023llama} provided three responses per data point, Zephyr only gave one. The advantage of multiple responses is the potential for higher accuracy. In cases of agreement, we considered the model confident and non-hallucinating. Contradictions suggested hallucination, but instead of simply discarding opinions, we devised rules to combine the responses.
\\
We implemented the judgment method with three configurations:
\\
Composition 1: Mistral and Zephyr commented, with Mistral judging based on their comments (label, score, description output).
\\
Composition 2: Same as 1, but Zephyr judged.
\\
Composition 3: Llama2 and Mistral commented, with Zephyr choosing whose opinion was more reliable. 
\\
\section{Experiments and Experimental Setup}

\subsection{Semantic Similarity}
In semantic similarity methods, we consider the target output in MT and DM tasks, and the input in PG tasks as reference. The similarity score between the reference and hypothesis for each data point is computed, and 1 minus this score is considered as the probability of hallucination. Probabilities below 0.5 are classified as "Not Hallucination", while others are labeled as "Hallucination".
\\
\\
Using the \textbf{LaBSE} model, we obtained embeddings of the reference and hypothesis, and the cosine similarity of these two embeddings was considered as the similarity score.
\\
\\
Using the prompt "Is the Sentence supported by the Context above?" we asked each \textbf{LLM} (Zephyr and Mistral) to provide a score between 1 and 5 determining the similarity of the reference and hypothesis. These values were then normalized to range between 0 and 1.
\\
\subsection{Natural Language Inference (NLI)}
For the NLI method, we used the reference as the premise and the hypothesis of each record as the hypothesis in the NLI model. This model outputs three probabilities which determine the probability of entailment, neutral, and contradiction between the two inputs. We utilized the probability of entailment for hallucination detection as it yielded better results compared to the other two options on the validation data.
\\
\subsection{Judgement method}
Commenting LLMs were prompted to check whether the sentence was supported by the context and were asked to return a label, a score between 1 and 5, and an explanation.
\\
\\
Prompt for commenting LLMs:
\\
\lightblacktext{Answer the following question using this JSON format: {{answer: (yes, no or maybe), score: (an integer number between 1 and 5, which 1 is for not supported and 5 is for fully supported), description: (a description for your answer)}}.\\
{question} Is the Sentence supported by the Context above?
}
\\
\\
Three outputs were taken from Llama2 and Mistral for each data, and these three outputs were converted into one with rules.
\\
In the first two compositions, Mistral and Zephyr were commenters and the judge was asked to return an output with the same label, score, and explanation format after reviewing their explanations.
To see the result of self-correction, the judge was selected from among the commentators. The first time was Mistral and the second time was Zephyr.
\\
This was the prompt:
\\
\lightblacktext{Two experts are asked whether the given sentence supports the given context or not. We received two responses from these two experts. According to the explanations of these two experts, what is your decision? return your response in this JSON format {{label: (yes/no), score: (an integer number between 1 and 5, which 1 is for not supported and 5 is for full supported), explanation: (text)}}.}
\\
\\
In the third combination, the commentators were Llama2 and Mistral, and Zephyr was the judge. This time, we changed the prompt to the judge so that he chose only one of the two opinions as the more correct opinion.
\\
This was the judge's prompt:
\\
\lightblacktext{Answer the following question.\\
{question}\\
I asked two experts to determine whether the Sentence is supported by the Context or not.\\
Above are their explanations.Now judge which one gave a better reason. Give me just the index of the best expert with no explanations using this JSON format: {{index: (an integer number between 0 and 1, which 0 is for the first, 1 is for the second)}}.
}
\\
\\
\section{Results and Analysis}
\begin{table*}
\centering
\begin{tabular}{lllll}
\toprule
\space & \textbf{Model Aware} &\space & \textbf{Model Agnostic} \\
\space & Accuracy & Correlation($\rho$) & Accuracy & Correlation($\rho$) \\
\midrule
LaBSE & 0.706 & 0.426 & 0.658 & 0.464\\
Zephyr & 0.700 & 0.370 & 0.694 & 0.423\\
Mistral & 0.630 & 0.213 & 0.568 & 0.183\\
\textbf{DeBERTa-v3(NLI Model)} & \textbf{0.777} & \textbf{0.661} & \textbf{0.780} & \textbf{0.689} \\ 
Mistral Judge & 0.644 & 0.291 & 0.610 & 0.250\\
(Zephyr \& Mistral Reasons)\\
Zephyr Judge & 0.686 & 0.352 & 0.692 & 0.405\\
(Zephyr \& Mistral Reasons)\\
Zephyr Judge & 0.624 & 0.293 & 0.548 & 0.249\\
(LlaMa2 \& Mistral Reasons)\\
\\\midrule
\end{tabular}

\caption{Experiment Results}
\label{tab:accents}
\end{table*}

\begin{figure}
    \centering
    \includegraphics[width=0.5\textwidth]{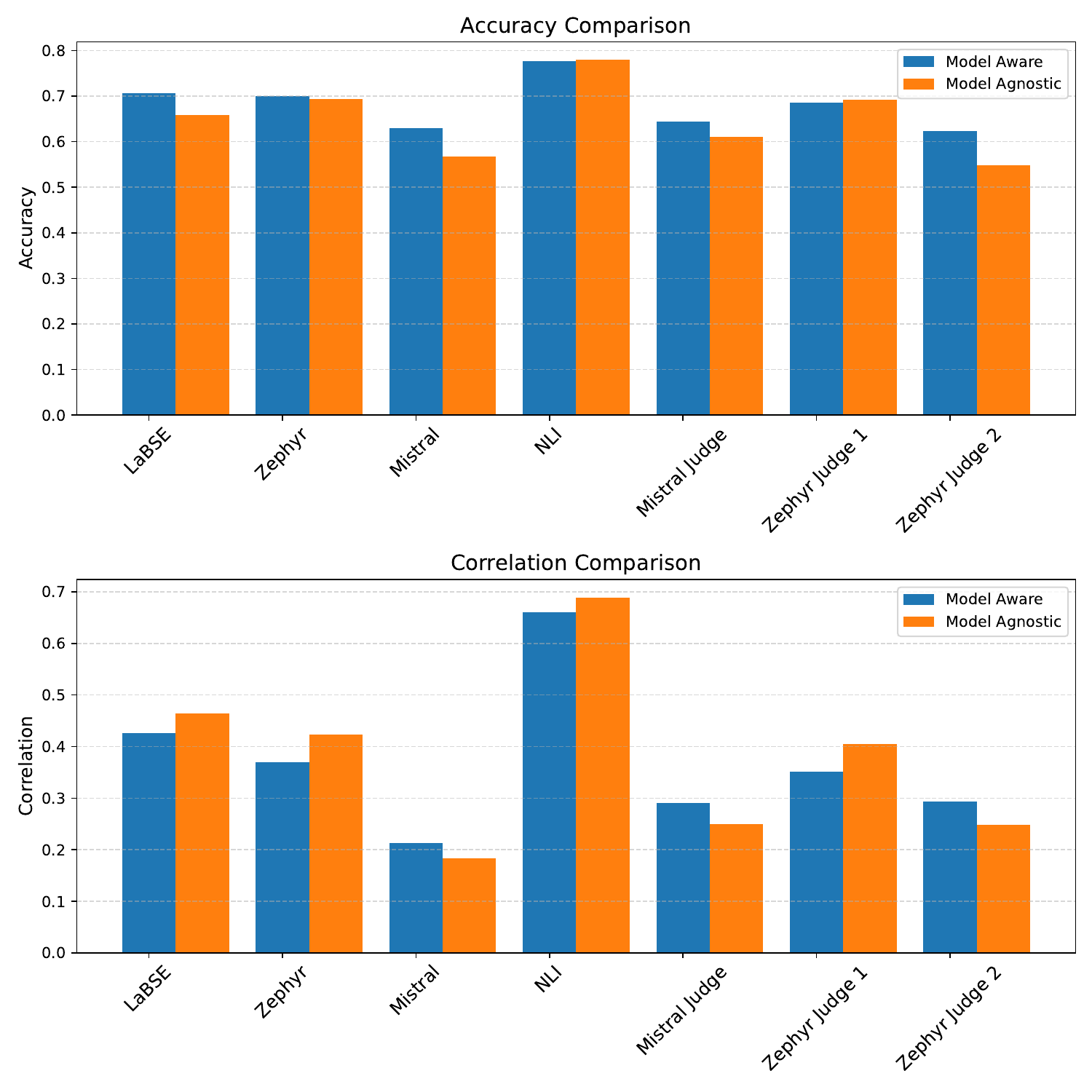} 
    \caption{Comparison of accuracy and correlation scores across multiple models in model-aware and model-agnostic datasets.}
    \label{fig:example}
\end{figure}

\textbf{Semantic Similarity and NLI:}
\\
Table \ref{tab:accents} and Figure \ref{fig:example} showcases the results of hallucination classification on trial data for each method we employed. Based on these findings, the Semantic Similarity method, utilizing models like LaBSE and Zephyr, demonstrates moderate accuracy and correlation scores. While LaBSE holds promise due to its renowned semantic similarity capabilities, there's room for improvement. 
\\
Notably, among the two Language Learning Models (LLMs) utilized in the semantic similarity approach, Zephyr yielded considerably better results than Mistral. This was also evident in the validation data, which influenced our decision to incorporate it into all our judgmental method experiments.
\\
\\
The DeBERTa model or NLI method outperforms all other methods, suggesting that incorporating natural language inference strengthens our ability to discern hallucinations by capturing semantic relationships between generated text and reference data. 
\\
\\
\\
\textbf{Ensemble LLMs and the Judgment Method:}
\\
\\
Furthermore, our study explored the effectiveness of Ensemble LLMs utilizing a Judgment Method. Surprisingly, the results indicate that Ensemble LLMs' performance wasn't superior to the previous method; in fact, it was even lower. Zephyr, acting as a judge, exhibited lower accuracy and correlation scores compared to Zephyr alone. While Mistral, in the role of a judge, showed improved performance compared to Mistral alone, it still falls short of methods like DeBERTa and the LaBSE model, suggesting limitations in this approach's effectiveness for hallucination detection.
\\
\section{Conclusion}
This study investigated three distinct methods for hallucination detection in language models: the Semantic Similarity method, NLI and the Ensemble LLMs with Judgment method. By analyzing and comparing these approaches, we gained valuable insights into their efficacy and suitability for identifying hallucinatory content in model-generated text.
\\
\\
\textbf{Semantic Similarity Method:}
\\
The utilization of pre-trained models such as LaBSE or large language models (LLMs) like Zephyr has demonstrated the potential for hallucination detection by assessing coherence between generated text and reference data through the Semantic Similarity method. Our findings underscore the effectiveness of employing specialized embedding models like LaBSE, which consists of approximately 500 million parameters, yielding comparable results to LLMs like Zephyr with 7 billion parameters. This highlights the efficiency of utilizing specialized embedding models for such tasks. However, while the semantic similarity method has shown moderate success, it falls short of being deemed the optimal choice for hallucination detection. Relying solely on similarity may not adequately capture all forms of hallucination and could prove insufficient across various tasks and scenarios. It's worth noting that exploring these limitations is beyond the scope of this paper and warrants further investigation by other researchers.
\\
\\
\textbf{NLI:}
\\
In conclusion, our findings underscore the efficacy of the NLI method as the optimal model for our study, indicating its potential utility in hallucination detection through entailment scoring. 
However, similar to the Semantic Similarity method, it is essential to acknowledge the inherent limitations in extrapolating the concept of entailment to the domain of hallucination detection. While NLI datasets offer valuable insights, they may not encompass the full complexity of hallucination phenomena. Therefore, while NLI tasks present promising avenues for further exploration in this area, additional research is warranted to ascertain their applicability and effectiveness in comprehensive hallucination detection frameworks.
\\
\\
\textbf{Ensemble LLMs with Judgment Method:}\\
This novel approach introduced multi-agent feedback and ensemble modeling for hallucination detection. LLMs acted as commentators, providing input to a "judge" model for final decision-making, aiming to enhance individual models' reasoning and decision-making. While not exceeding initial expectations, our experiments yielded valuable insights into the ensemble's effectiveness, with varying accuracy and correlation depending on composition and judging strategies.
\\
\\
\textbf{Discussion and Future Directions:}
\\
Although the performance of the Ensemble LLMs with Judgment method wasn't as promising as envisioned, it sheds light on the complexities of hallucination detection and the limitations of current methods. One of the key challenges in these methods is finding the optimal prompt to detect hallucinations in the language model, and the utilization of prompt engineering methods can be beneficial in this regard. The potential for improved results using larger, more capable LLMs suggests avenues for future exploration.
\\
\\
Overall, this study contributes to addressing challenges posed by hallucinations in language models. By evaluating and comparing distinct detection methodologies, we highlight the strengths and weaknesses of each approach, paving the way for future research and development in this crucial area.


\bibliography{anthology,custom}

\end{document}